\documentclass[letterpaper, 10 pt, conference]{ieeeconf}  

\IEEEoverridecommandlockouts                              
\usepackage{amsmath} 

\title{\LARGE \bf
MGMD-GAN: Generalization Improvement of Generative Adversarial Networks with Multiple Generator Multiple Discriminator Framework Against Membership Inference Attacks
}

\author{Nirob Arefin \\
        Department of Computer Science and Engineering \\
        Bangladesh University of Engineering and Technology \\
        1505050.na@ugrad.cse.buet.ac.bd
}

\usepackage[T1]{fontenc}
\usepackage{graphicx}

\begin{document}

\maketitle
\thispagestyle{empty}
\pagestyle{empty}

\begin{abstract}

Generative Adversarial Networks (GAN) are among the widely used Generative models in various applications. However the original GAN architecture may memorize the distribution of the training data and, therefore, poses a threat to Membership Inference Attacks. In this work, we propose a new GAN framework that consists of Multiple Generators and Multiple Discriminators (MGMD-GAN). Disjoint partitions of the training data are used to train this model and it learns the mixture distribution of all the training data partitions. In this way, our proposed model reduces the generalization gap which makes our MGMD-GAN less vulnerable to Membership Inference Attacks. We provide an experimental analysis of our model and also a comparison with other GAN frameworks.  

\end{abstract}
\section{INTRODUCTION}

Improvement of the machine learning models for various applications greatly depends on the amount of data they can be provided during training. Although the research community shares datasets among themselves, there are certain areas like medical records and other sensitive information for which publicly available data are so limited. To overcome this limitation and protect the privacy of the original data, various generative models have been proposed in the literature that can generate fake examples from a small set of data.

Generative Adversarial Network (GAN) [1] is a widely used generative model that can produce much more promising fake samples. However, recent studies found that the classical GAN model does not generalize well, and is thus vulnerable to Membership Inference Attacks [2] where an attacker can infer whether a particular sample is used to train a model. 

It is a well-known intuition in the literature that reducing the generalization gap and protecting an individual's privacy share the same goal of encouraging a neural network to learn the population's features instead of memorizing the features of each individual [3]. This implies that the lesser amount of training data leakage possibility can be ensured by the smaller generalization gap.

In this paper, we focus on minimizing the Membership Attack threats by reducing the generalization gap. We propose a Multiple Generators Multiple Discriminators (MGMD-GAN) framework where the training dataset is divided into \textit{K} disjoint partitions. Each Generator Discriminator pair is trained over a single partition.

In recent years, several GAN frameworks have been proposed which consist of multiple generators and discriminators like privGAN [4] and MIX-GAN [5]. However, our model differs from them as we do not consider a built-in adversary like in privGAN and our model works on disjoint training data partitions.

The rest of the paper is structured as follows. We briefly describe the related works in Section II. In Section III, we introduce the preliminaries. In Section IV, we present the MGMD-GAN and its theoretical properties. We present the evaluation results of our framework in Section V. Finally, we conclude the paper in Section VI. 

\section{RELATED WORKS}

Since the introduction of Generative Adversarial Networks [1], a wide variety of GAN frameworks have been proposed considering different aspects. In this section, we describe a brief literature review of those works that focus on generalization and privacy concerns and use more than one generator and/or discriminator.

Authors in [5] proposed MIX-GAN where they first proved that multiple generators and multiple discriminators GANs can improve the chance of getting approximate pure equilibrium. Along with multiple generator-discriminator pairs, a built-in adversary was also proposed in Priv-GAN [4] where the goal of the built-in adversary, the privacy discriminator, is to prevent the generators from memorizing their corresponding data splits. Liyang et al. [6], based on differential privacy [7], proposed DPGAN which applies a combination of noise and gradient clipping on the weights only. Authors in [8] proposed a framework MDGAN with multiple discriminators and a single generator with distributed settings. In PAR-GAN [9] a single generator and multiple discriminator model is proposed but with centralized settings. PAR-GAN also focuses on mainly reducing the generalization gap by approximating a mixture distribution of all the data partitions it creates during training. It is also shown that PAR-GAN can outperform existing classical GAN models in respect of reducing generalization gaps but the number of data partitions needs to be determined empirically. Authors in [10] proposed a different privacy-preserving framework where they manipulate the original data instead of adding noise and use a variational autoencoder (VAE) model to construct the synthetic data.

\section{PRELIMINARIES}
\subsection{Generative Adversarial Networks}
In Generative Adversarial Networks, there exist two different neural networks and they are being trained simultaneously - a generative model \textit{G} and a discriminative model \textit{D}. The Generator \textit{G} tries to generate fake samples, by mapping random noise to the training distribution, to make the Discriminator fool. The goal of the Discriminator \textit{D} is to correctly label a sample data whether it is coming from the Generator (fake) or from the actual training samples (real). 

Let's consider the input noise distribution of the generator is \(p_z (z)\) and the distribution of the real samples is \( p_{data} (x) \). In GAN, the Generator \textit{G} and the discriminator \textit{D} play a min-max game until they reach the Nash equilibrium [5] by optimizing the following value function -
\begin{equation} \label{eq1}
    \begin{split}
        \min_G \max_D V(G,D) = E_{x \sim p_{data} (x) } [\phi(D(x))] \\ 
        +   E_{z \sim p_{z} (z) } [\phi(1 - D(G(z)))]
    \end{split}
\end{equation}

Here, \(\phi\) is called a measuring function. In classic GAN [1], \(\phi(x) = log(x) \) was used. Another popular measuring function for GAN is the Wasserstein distance where \(\phi(x) = x \) is used. Martin et al. [11] proposed and showed that WGAN can improve performance.

\subsection{Membership Inference Attack}
In MIA, the adversary trains a separate attack model that takes some features of the target model as input. The goal of this attack model is to predict whether a particular sample was part of the real training samples. Specifically, for a Generator model \textit{G}, a Discriminator model \textit{D} and a training dataset \textit{X}, the attack model tries to learn the function \textit{f(G,D,x)} where: 
\begin{equation} \label{eq2}
    \Pr (x \in X) = f(G,D,x)
\end{equation}
The attack model can target the generator and/or the discriminator.

The MIA attacks against GAN can be classified into some types. Out of which our focus in this work is based on the \textit{White-Box} scenario where the adversary can gain access to the model parameters.

\section{METHODOLOGY}
In this section, we describe our proposed MGMD-GAN framework with a mathematical formulation. We also describe if our proposed framework is able to defend against Membership Inference Attacks.

\subsection{MGMD-GAN Framework}

\begin{figure}[h]
    \centering
    \includegraphics[width=0.9\columnwidth]{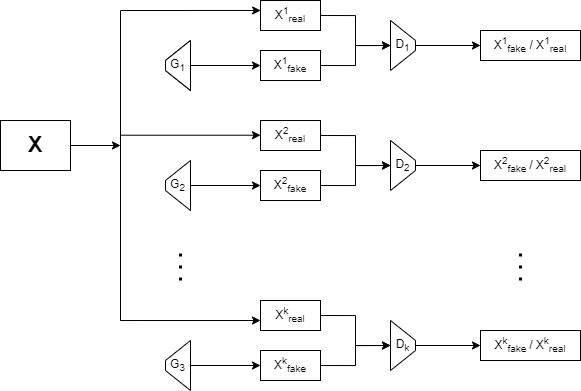}
    \caption{MGMD-GAN Architecture Overview}
    \label{fig:fig1}
\end{figure}

The architecture of MGMD-GAN is presented in Figure \ref{fig:fig1}. The training dataset is divided into \textit{K} disjoint partitions and then fed into \textit{K} Generator-Discriminator pairs. Each Discriminator here needs to identify whether a sample belongs to a particular partition or not, along with whether it is real or fake. Each Generator here is trained with respect to the corresponding discriminator.

\subsection{Mathematical Formulation of MGMD-GAN}
Let's assume a training dataset \textit{X} with distribution \(p_x\). For our framework, this training dataset is divided into \textit{k} disjoint partitions \(X_{real}^1\), \(X_{real}^2\), \(X_{real}^3\), ..., \(X_{real}^k\) . Let's assume that the distributions of these data partitions are \(p_{x_1}\), \(p_{x_2}\), \(p_{x_3}\), ..., \(p_{x_k}\), respectively. Each Generator-Discriminator pair is trained on separate data partitions and competes with each other. Therefore, the value function for our MGMD-GAN can be described as - 

\begin{equation} \label{eq3}
    \begin{split}
        \min_{G_{i=1}^k} \max_{D_{i=1}^k} V(G,D) = E_{x \sim p_{x_i} (x) } [\phi(D_i(x))] \\ 
        +   E_{z \sim p_{z} (z) } [\phi(1 - D_i(G_i(z)))]
    \end{split}
\end{equation}

The Loss function for each Discriminator and Generator can be described as the following respectively - 

\begin{equation} \label{eq4}
    \begin{split}
        L_{D_i} = E_{x \sim p_{x_i} (x) } [\phi(D_i(x))] 
        +   E_{z \sim p_{z} (z) } [\phi(1 - D_i(G_i(z)))]
    \end{split}
\end{equation}

\begin{equation} \label{eq5}
    \begin{split}
        L_{G_i} = \frac{1}{k} \sum_{i=1}^k E_{z \sim p_{z} (z) } [\phi(1 - D_i(G_i(z)))]
    \end{split}
\end{equation}

\section{EXPERIMENT}
In this section, we present details about our experiments. For our experiments, we took PAR-GAN [9] as a baseline GAN model. We also took help from their implementation source code [12].

\subsection{Experimental Setup}
We evaluated our model on the widely used MNIST [13] dataset which is a handwritten image dataset. The training set of MNIST has 60,000 samples and the test set has 10,000.

We developed our framework using Tensorflow version 2.6.0 [14]. The generator and the discriminator model of our framework are conventional neural networks and follow the model architecture as described in [9]. We run our MGMD-GAN framework on a local machine. In our evaluation, we considered two types of objective functions - Wasserstein Distance ( \(\phi(x) = x \) ) and JS divergence ( \(\phi(x) = log(x) \) ). We compare the performance of our MGMD-GAN with the PAR-GAN framework. For this comparison, we configured the PAR-GAN in our local machine. In our experiment, all models are trained for 1500 epochs with a batch size of 64.

\subsection{Comparison of Generalization Gap}

We can conjecture how well a GAN performs by visually comparing the distribution discriminators' prediction scores for the training data with the distribution on the holdout data. It becomes harder for an adversary to tell whether a sample belongs to actual training data or not if the distributions are more similar. In our MGMD-GAN architecture, multiple discriminators are present. Therefore, we merge all discriminators' predicted scores into one distribution. 

\begin{figure}[h]
    \centering
    \includegraphics[width=\columnwidth]{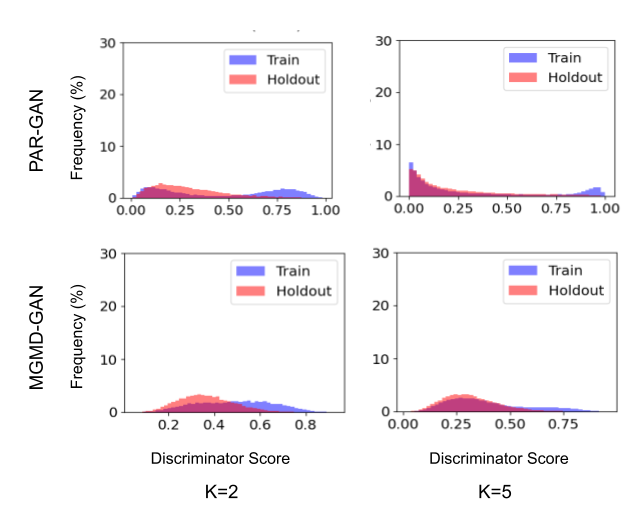}
    \caption{Comparison of prediction scores by discriminators when JS divergence is used as value function}
    \label{fig:fig2}
\end{figure}

In figure \ref{fig:fig2}, a comparison of discriminators' prediction scores between PAR-GAN and MGMD-GAN is presented when JS divergence is used as a value function. We can observe that for \textit{k=2}, PAR-GAN predicts most of the training data with 1.0 and most of the holdout data with 0.0. There also presents a notable generalization gap between the distributions of training and holdout data. On the other hand, our proposed MGMD-GAN predicts most of the samples around 0.5 which means it reduces the overfitting better than the PAR-GAN. If the number of data partitions is increased ( \textit{k=5} ), PAR-GAN does not improve much. The generalization gap between training and holdout data distribution does not become smaller. Like the previous scenario, training data and holdout data are predicted mostly with 1.0 and 0.0 respectively. On the other hand, MGMD-GAN shows better performance when \textit{k=5} partitions are used. The generalization gap becomes smaller than the scenario with \textit{k=2}. It also predicts most data around the center.

\begin{figure}[h]
    \centering
    \includegraphics[width=\columnwidth]{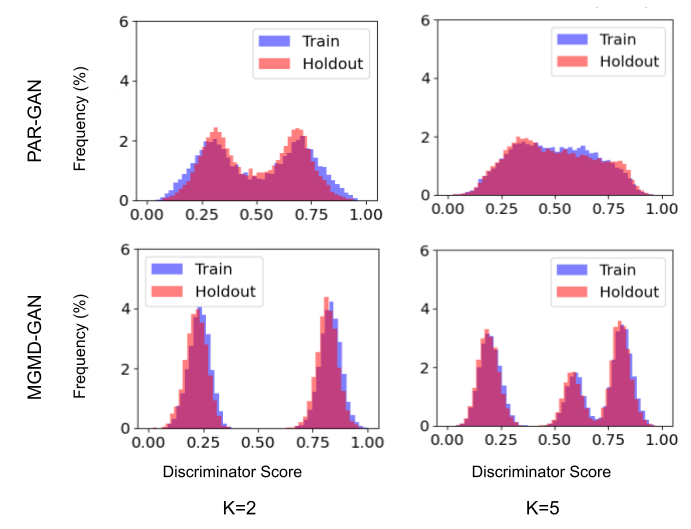}
    \caption{Comparison of prediction scores by discriminators when Wasserstein distance is used as value function}
    \label{fig:fig3}
\end{figure}

If we consider Wasserstein distance as a value function,  discriminators' prediction scores show different kinds of distributions which are presented in figure \ref{fig:fig3}. But we can observe that even in this case, our proposed MGMD-GAN shows a more similar distribution between training and holdout data than PAR-GAN. 

\subsection{ Attack on Discriminators}

In our experiment of MIA evaluation, we have set the training and holdout data to be of the same size. Therefore, the accuracy of MIA upon random guess is around 0.5.

\begin{table}[h]
\begin{tabular}{|l|l|l|}
\hline
Methods & JS-Divergence & \begin{tabular}[c]{@{}l@{}}Wasserstein \\ Distance\end{tabular} \\ \hline
PAR-GAN (k=2)  & 0.8436          & 0.5723          \\ \hline
PAR-GAN (k=5)  & 0.7181          & \textbf{0.5591} \\ \hline
MGMD-GAN (k=2) & 0.7248          & 0.5647          \\ \hline
MGMD-GAN (k=5) & \textbf{0.6728} & 0.561           \\ \hline
\end{tabular}
\caption{MIA attack accuracy on discriminators}
\label{tab:my-table1}
\end{table}

Table \ref{tab:my-table1} shows the white-box attack accuracy on discriminators for various methods of PAR-GAN and MGMD-GAN. For JS-Divergence value function, we can observe that MGMD with \textit{k=5} partitions performs better than other methods. Although for Wasserstein distance PAR-GAN with \textit{k=5} performs better but our MGMD-GAN does not perform much worse than PAR-GAN.   

\subsection{ Attack on Generators}

\begin{table}[h]
\begin{tabular}{|l|l|l|}
\hline
Methods & JS-Divergence & \begin{tabular}[c]{@{}l@{}}Wasserstein \\ Distance\end{tabular} \\ \hline
PAR-GAN (k=2)  & 0.8           & \textbf{0.66} \\ \hline
PAR-GAN (k=5)  & 0.66          & 0.72          \\ \hline
MGMD-GAN (k=2) & \textbf{0.65} & 0.69          \\ \hline
MGMD-GAN (k=5) & 0.692         & 0.708         \\ \hline
\end{tabular}
\caption{MIA attack accuracy on generators}
\label{tab:my-table2}
\end{table}

The accuracy of MIA on generators for various models is presented in Table \ref{tab:my-table2}. We notice that MGMD-GAN performs better with \textbf{k=2} when JS-Divergence is used as a value function. However, when Wasserstein distance is considered, PAR-GAN with \textit{k=2} performs better. MGMD-GAN also performs close to PAR-GAN in the case of the Wasserstein distance value function when \textit{k=2} is considered.

\section{CONCLUSION}

In this paper, we proposed a new GAN framework, MGMD-GAN, which focuses on reducing the generalization gap in order to make the GAN model protective of Membership Inference Attacks. We analytically showed that our proposed model can mitigate the overfitting problem thus ensuring better generalization. We also presented experimental results on the MNIST dataset which show that our MGMD-GAN does reduce the generalization gap. Comparison with a state-of-the-art GAN model shows that MGMD-GAN can reduce the MIA attack accuracy both on generators and discriminators if the number of data partitions is carefully chosen.

In the future, we plan to use our framework with other structured and non-structured datasets so that we can identify the proper relation of our method with the distribution of the training data samples. 


\end{document}